%% file: acl2020.tex
\documentclass[11pt,a4paper]{article}

\usepackage[hyperref]{acl2020}
\usepackage{times}
\usepackage{latexsym}
\usepackage{color}
\usepackage{mciteplus}
\usepackage{graphicx}
\usepackage{amsmath}
\usepackage{amsthm, amssymb, bm}
\usepackage{tabulary}
\usepackage{booktabs}
\usepackage{multirow}
\usepackage{booktabs}
\usepackage{soul}
\usepackage{hyperref}
\usepackage{bbold}
\usepackage{subfigure}

\usepackage[ruled,linesnumbered]{algorithm2e}

\usepackage{microtype}

\aclfinalcopy 

\usepackage{pifont}

\newcommand{\F}{Figure}

\newcommand{\T}{Table}
\renewcommand{\S}{Sec.}
\newcommand{\A}{Alg.}

\newcommand{\ignore}[1]{}

\newcommand{\teql}{\textsc{MT-Teql}}

\makeatletter
\newcommand{\thickhline}{%
    \noalign {\ifnum 0=`}\fi \hrule height 1pt
    \futurelet \reserved@a \@xhline
}
\newcolumntype{"}{@{\hskip\tabcolsep\vrule width 1pt\hskip\tabcolsep}}
\makeatother

\title{\teql: Evaluating and Augmenting Consistency of Text-to-SQL Models with
  Metamorphic Testing}

\author{Pingchuan Ma \\
  Hong Kong University of Science \\and Technology\\
  \texttt{pmaab@cse.ust.hk} \\\And 
  Shuai Wang \\
  Hong Kong University of Science \\and Technology\\
  \texttt{shuaiw@cse.ust.hk} \\}

\date{}

\begin{document}
\maketitle
\begin{abstract}
Text-to-SQL is a task to generate SQL queries from human utterances. However,
due to the variation of natural language, two semantically equivalent utterances
may appear differently in the lexical level. Likewise, user preferences (e.g., 
the choice of normal forms) can lead to dramatic changes in table structures
when expressing conceptually identical schemas. Envisioning the general
difficulty for text-to-SQL models to preserve prediction consistency against
linguistic and schema variations, we propose \teql, a \textbf{M}etamorphic
\textbf{T}esting-based framework for systematically evaluating and augmenting the
consistency of \textbf{TE}xt-to-S\textbf{QL} models. Inspired by the principles
of software metamorphic testing, \teql\ delivers a model-agnostic framework which
implements a comprehensive set of metamorphic relations (MRs) to conduct
semantics-preserving transformations toward utterances and schemas. Model
Inconsistency can be exposed when the original and transformed inputs induce
different SQL queries. In addition, we leverage the transformed inputs to
retrain models for further model robustness boost. Our experiments show that our
framework exposes thousands of prediction errors from SOTA models and enriches
existing datasets by order of magnitude, eliminating over 40\% inconsistency
errors without compromising standard accuracy.
\end{abstract} 

\input{1-intro.tex}

\input{2-teql.tex}

\input{3-testing.tex}

\input{4-augmenting.tex}



\input{7-related.tex}

\section{Conclusion \& Availability}
We have presented \teql, a metamorphic testing framework conducting
model-agnostic testing on the consistency and coherence of text-to-SQL models.
Our evaluation shows that de facto models, despite its promising performance in
standard benchmark accuracy, manifest considerable inconsistency errors with
respect to either utterances or schema variants. We further propose data
augmentation strategies to extend the standard benchmark set and train
text-to-SQL models with much higher consistency and also retaining comparable
benchmark accuracy. 
We commit to make \teql\ publicly available, including all the code, synthetic
data and augmented models. We will maintain \teql\ to benefit follow-up
research.


\bibliography{acl2020}
\bibliographystyle{acl_natbib}

\appendix

\end{document}

%% file: 1-intro.tex
\section{Introduction}
\label{sec:introduction}
\begin{figure}[!htbp]
\centering
\includegraphics[width=0.9\columnwidth]{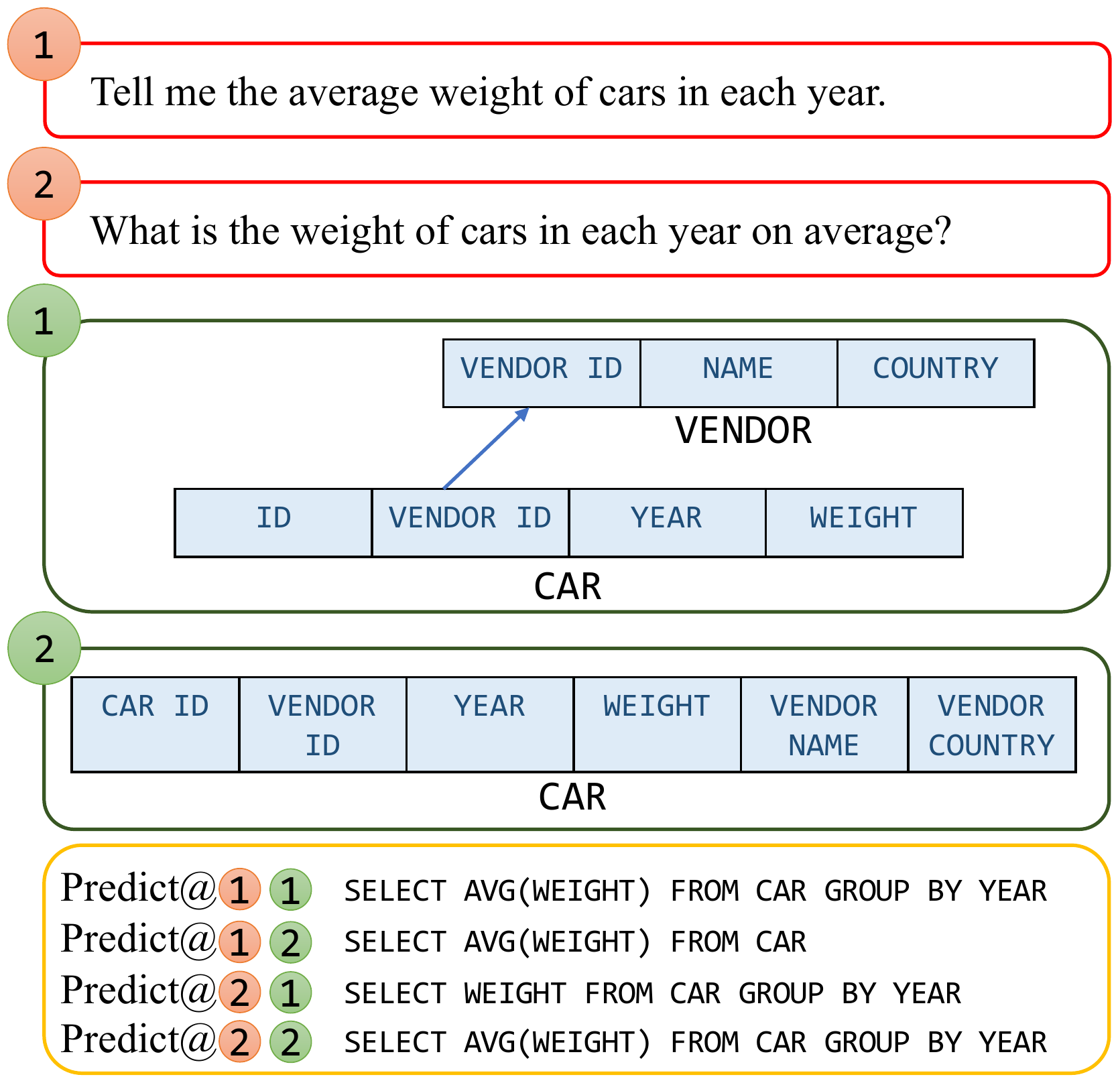}
\caption{Inconsistent predictions on semantically equivalent inputs.}
\label{fig:example}
\end{figure}

Text-to-SQL is a task that translates natural language statements to SQL
queries, which is expected to serve as a handy interface where even non-expert
users can retrieve data from database. Recent advances show potentials of neural
networks in synthesizing nested cross-domain SQL
queries~\cite{guo2019towards,wang2019rat,choi2020ryansql}.
Despite the spectacular progress, however, text-to-SQL models still face
considerable challenges due to the high flexibility of natural language
utterances and database schema design. One user intent can be expressed in
multiple lexically different utterances. Similarly, one conceptual model (in the
form of entity-relationship diagram) can be implemented in multiple
structurally-different schemas. Consider the erroneous predictions given in
\F~\ref{fig:example}, where given \textit{semantically equivalent} utterances or
schemas in different forms, text-to-SQL models yield inconsistent outputs.

These inconsistencies exhibit weakness of current models on generalization
ability and robustness. Standard benchmarks, e.g., Spider~\cite{Spider}, are
usually crafted following the best practice in schema design. In other words,
standard benchmarks, as will be shown in our findings (see
\S~\ref{subsec:findings}), are not sufficient to assess models under real-life
noisy scenarios. For instance, users may denormalize schema for online
analytical applications or may leverage an inadequate albeit functional database
schema (e.g., lacking of foreign key constraints). Such variations, together
with insufficient metrics to assess the inconsistency of text-to-SQL models,
likely become a practical challenge to impede the adoption of text-to-SQL models
in real-world scenarios.

With recent progresses in testing NLP
models~\cite{ribeirobeyond,ma2020metamorphic,soremekun2020astraea},
testing-based approaches become popular to assess ``in the wild'' model
robustness in addition to standard metrics on hold-out
accuracy~\cite{ribeirobeyond}.
%
However, we note that adapting existing testing-based methods to assess
text-to-SQL models are unattainable, whose reasons are three-fold. First,
typical text-to-SQL models employ a unique learning paradigm that jointly learns
the representations of utterances and database schemas in a common feature
space. However, previous testing approaches are \textit{not} designed to
transform schemas, thus presumably neglecting inconsistency defects due to
schema variations. Second, the problem of text-to-SQL is generally ill-posed, in
the sense that some text inputs are not translatable into SQL queries. Some
existing methods, by randomly substituting certain words with synonyms or typos,
can likely induce such ``unnatural'' inputs which are undesired in testing
text-to-SQL models~\cite{zeng2020photon}. Third, synthesizing text inputs based
on hand-coded templates~\cite{soremekun2020astraea,ribeirobeyond}, while
producing generally translatable inputs, is difficult to cover nested SQL
queries which are commonly seen in real-life usages. Given that SOTA models
usually have very high performance on simple queries, e.g.,
RAT-SQL~\cite{wang2019rat} has over 80\% accuracy on ``easy'' questions but less
than 40\% accuracy on ``extra hard'' questions, we see a demanding need to
synthesize more comprehensive utterances rather than only simple cases.
%

To address above challenges, we propose \teql, a metamorphic testing-based
framework for evaluating and augmenting text-to-SQL models. \teql\ transforms
seed inputs (utterance-schema pairs) via a comprehensive set of metamorphic
relations (MRs), where each MR specifies a \textit{semantics-preserving}
transformation scheme toward either utterances or schemas. By comparing SQL
queries derived from the original and transformed inputs, we assess the
consistency and coherence of the tested models.
%
From 1,034 testing samples in the Spider dev set~\cite{Spider},
\teql\ automatically generates a total of 62,430 transformed testing samples. We
test the consistency of popular text-to-SQL models, including the SOTA
RAT-SQL~\cite{wang2019rat}, with the transformed samples and identify
considerable inconsistency errors. We further augment models by extending the
training dataset with those transformed inputs into a synthetic dataset. We show
that training the text-to-SQL models with this dataset can effectively eliminate
over 40\% inconsistency errors while retaining comparable benchmark accuracy.

\smallskip
\noindent \textbf{Key Contributions.}~1) We propose \teql, a model-agnostic
framework to test the consistency of text-to-SQL models without a requirement
for manually labeled answers. 2) We design a comprehensive set of metamorphic
relations (MRs) to conduct semantics-preserving transformations on utterances
and schemas and expose inconsistency errors. 3) We propose three augmentation
schemes by leveraging the transformed inputs to improve model consistency. 4)
Our evaluation exposes considerable inconsistency defects, achieves effective
model consistency augmentation, and also reveals insightful observations on de
facto models that can be used for engineers to diagnose and improve models in
real-life usage.

%% file: 2-teql.tex
\section{Metamorphic Testing (MT)}
\label{sec:background}
Determining the correctness of SQL queries generated by text-to-SQL models for
arbitrary utterance-and-schema pairs generally requires human annotated ground
truth. In contrast, MT specifies to benchmark testing targets via
\textbf{metamorphic relations} (\textbf{MRs}) without the need of ground truth.
Each MR denotes a general and usually \textit{invariant property} of the testing
targets. For instance, to test the implementation of $sin(x)$, instead of
knowing the expected output of arbitrary floating-point input $x$ (which
requires considerable manual efforts), we assert that the MR $sin(x) = sin(\pi -
x)$ always holds when arbitrarily mutating $x$. A bug in $sin(x)$ is detected
when input $x$ and its mutation $(\pi - x)$ induce inconsistent outputs. To
date, MT has achieved major success in detecting bugs in software and AI
systems~\cite{chen1998metamorphic,segura2016survey}. This research further
leverages MT to evaluate the consistency of text-to-SQL models, by defining a
comprehensive set of MRs to conduct semantics-preserving transformations toward
natural language utterances and database schemas.


\begin{algorithm}[!htbp]
\caption{\teql}
\label{alg:teql}

\KwIn{model $\mathcal{M}:\mathcal{U}\times \mathcal{S}\to \mathcal{Q}$, 
seed utterance $u_0$, seed schema $s_0$}
\KwOut{inconsistent errors $E=\{(u_1,s_1),\cdots\}$}
$T \leftarrow \mathcal{MR}_1(u_0, s_0)\cup\cdots\cup\mathcal{MR}_n(u_0, s_0)$\;
$E \leftarrow \emptyset$\;
\ForEach{$(u',s') \in T$}{
    \If{$\mathcal{M}(u',s') \neq \mathcal{M}(u_0,s_0)$}{ 
        $E = E\cup \{(u',s')\}$
    }
}
\Return $E$\;
\end{algorithm}

\section{\teql}
\label{sec:teql}

\A~\ref{alg:teql} depicts the overall workflow of \teql, where given seed
utterance $u_0$, seed schema $s_0$, and model $\mathcal{M}$, \teql\ first
generates a collection $T$ of transformed utterances and schemas based on a set
of MRs (line 1). Then, for each pair of transformed utterances and schemas
$(u',s')$, \teql\ checks the consistency between SQL queries derived from
$(u',s')$ and queries derived from the original seed input $(u,s)$ (line 4).
Inconsistency-triggering inputs will be collected for model augmentation (line
5).

Note that standard benchmark accuracy is generally measured by comparing
prediction $\mathcal{M}(u',s')$ with human annotated ground truth. In contrast,
inspired by software metamorphic testing, \teql\ directly compares the
\textit{consistency} between $\mathcal{M}(u,s)$ and $\mathcal{M}(u',s')$ (line
4), thus alleviating manual efforts. In addition, considering text-to-SQL models
typically develop a \textit{joint} understanding of natural-language utterances
and schemas, \teql\ transforms both utterances and database schemas with a set
of semantics-preserving MRs (line 1), while existing general-purpose NLP model
testing works can likely neglect certain model inconsistencies as they only
focus on text~\cite{ribeirobeyond}.


\subsection{Metamorphic Relations (MRs)}
\label{subsec:mutation}

\input{tab/mutation-strategies.tex}


In contrast with previous works, \teql\ does \textit{not} mutate inputs from an
adversarial perspective (e.g., substitute words with typos or unconstrained
synonyms), nor does \teql\ break utterances/schemas into ``untranslatable''
forms to stress text-to-SQL models~\cite{zeng2020photon}. In fact, one key
strength of \teql\ is to perform semantics-preserving transformation and check
the logic consistency of generated SQL queries. This way, we alleviate the
necessity of manually checking the correctness of SQL queries, and the entire
workflow can be conducted fully automatically. As listed in
\T~\ref{tab:strategies}, we instantiate a total of 12 MRs to systematically
explore model defects. Following, we elaborate on the details of each MR.

\subsubsection{Utterance Transformation}
\label{subsec:utterance}
Usually, an utterance in the text-to-SQL task starts with a prefix (e.g., ``what
is'' and ``tell me'') and is followed by the real query body. We categorize 
frequently-used prefixes in utterances in \T~\ref{tab:prefix}. Here, common prefixes
do not contain the user intents while special prefixes can implicitly indicate 
some user intents (e.g., desired columns or aggregate functions). Observing that the
choice or even the occurrence of common prefixes usually does not affect the
user intent, we design three MRs focusing on transforming prefixes as follows.

\begin{table}[!ht]
  \centering
  \small
		\begin{tabular}{l|l}
      \hline
			\textbf{Type} & \textbf{Illustrative Examples}\\
      \hline
        Common Interrogative Prefix& what is/are, which is/are\\
        \hline
        Common Declarative Prefix & tell me, return, find, list\\
        \hline
        Special Interrogative Prefix & when, where, how many \\
        \hline
        Special Declarative Prefix & count \\
      \hline
		\end{tabular}
  \caption{Categorizations of frequently-used prefixes in typical text-to-SQL 
  utterances.}
  \label{tab:prefix}
\end{table}

\smallskip
\noindent \textbf{Prefix Insertion (PI).}~In case the utterance starts from an
explicit interrogative (either common or special) prefix, our first MR specifies 
inserting a common declarative prefix at the beginning of the sentence. For example, 
``\emph{tell me} what is the age of all singers?''

\smallskip
\noindent \textbf{Prefix Removal (PR).}~As mentioned, the occurrence of prefixes
does not affect user intents. Hence, our second MR specifies removing the common 
prefixes and feeding the query body to the model. For example, ``\st{\emph{what is}} 
the age of all singers?''

\smallskip
\noindent \textbf{Prefix Substitution (PS).}~Likewise, the choice of common prefixes
before query body also does not affect user intents. Hence, we further define an
MR to replace the prefix in an utterance with another prefix, e.g., ``\emph{tell
  me} \st{\emph{what is}} the age of all singers?''

\smallskip
\noindent \textbf{Synonym Substitution (SS).}~In addition to transforming
prefixes, we further propose an MR focusing on replacing certain tokens in an
utterance with their synonyms. While such ``synonym transformation'' scheme has
been proposed in previous general-purpose NLP model testing
frameworks~\cite{ribeirobeyond}, \T~\ref{tab:mapping} incorporates domain
knowledge to manipulate a collection of specific synonyms tailored for the SQL
language.

\begin{table}[!ht]
  \centering
  \small
		\begin{tabular}{l|c}
      \hline
			\textbf{AGG} & \textbf{Textual Description}\\
      \hline
        \texttt{MIN} & minimal, minimum, lowest, smallest\\
        \hline
        \texttt{MAX} & maximal, maximum, highest, largest\\
        \hline
        \texttt{COUNT} & the (total) \{number, count, amount\} of \\
        \hline
        \texttt{SUM} & the (total) \{sum, amount\} of \\
      \hline
      \texttt{AVG} & the \{mean, average\} of\\
      \hline
		\end{tabular}
  \caption{Illustrative mapping relations between aggregate functions
    (\textbf{AGG}) and textual descriptions.}
  \label{tab:mapping}
\end{table}

We list the synonyms mapping rules in \T~\ref{tab:mapping}. Overall, one
aggregate function in the SQL language can be expressed in multiple ways. For
instance, the last row in \T~\ref{tab:mapping} specifies that ``the \emph{mean}
of'' can be replaced into ``the \emph{average} of'' while retaining the derived
aggregate function \texttt{AVG}. Hence, by replacing tokens grouped together
(e.g., ``minimal'' $\rightarrow$ ``minimum'') in \T~\ref{tab:mapping}, the
derived SQL queries should preserve the same aggregate function, e.g.,
\texttt{MIN}.

Note that different aggregate functions may also be expressed in the same form.
For example, ``the \emph{amount} of'' can indicate either \texttt{SUM} or
\texttt{COUNT} aggregates in SQL (see the 3rd and 4th rows in
\T~\ref{tab:mapping}).
It is generally more challenging for models to infer an implicit aggregate
function based on the context if we substitute ``the \emph{sum} of'' to ``the
\emph{amount} of''; however, as such opaque aggregate function indicators widely
exist in the wild, we deem it necessary to stress models in this way.

\subsubsection{Schema Transformation}
\label{subsubsec:schema}

\begin{figure}[!htbp]
\centering
\includegraphics[width=1.01\columnwidth]{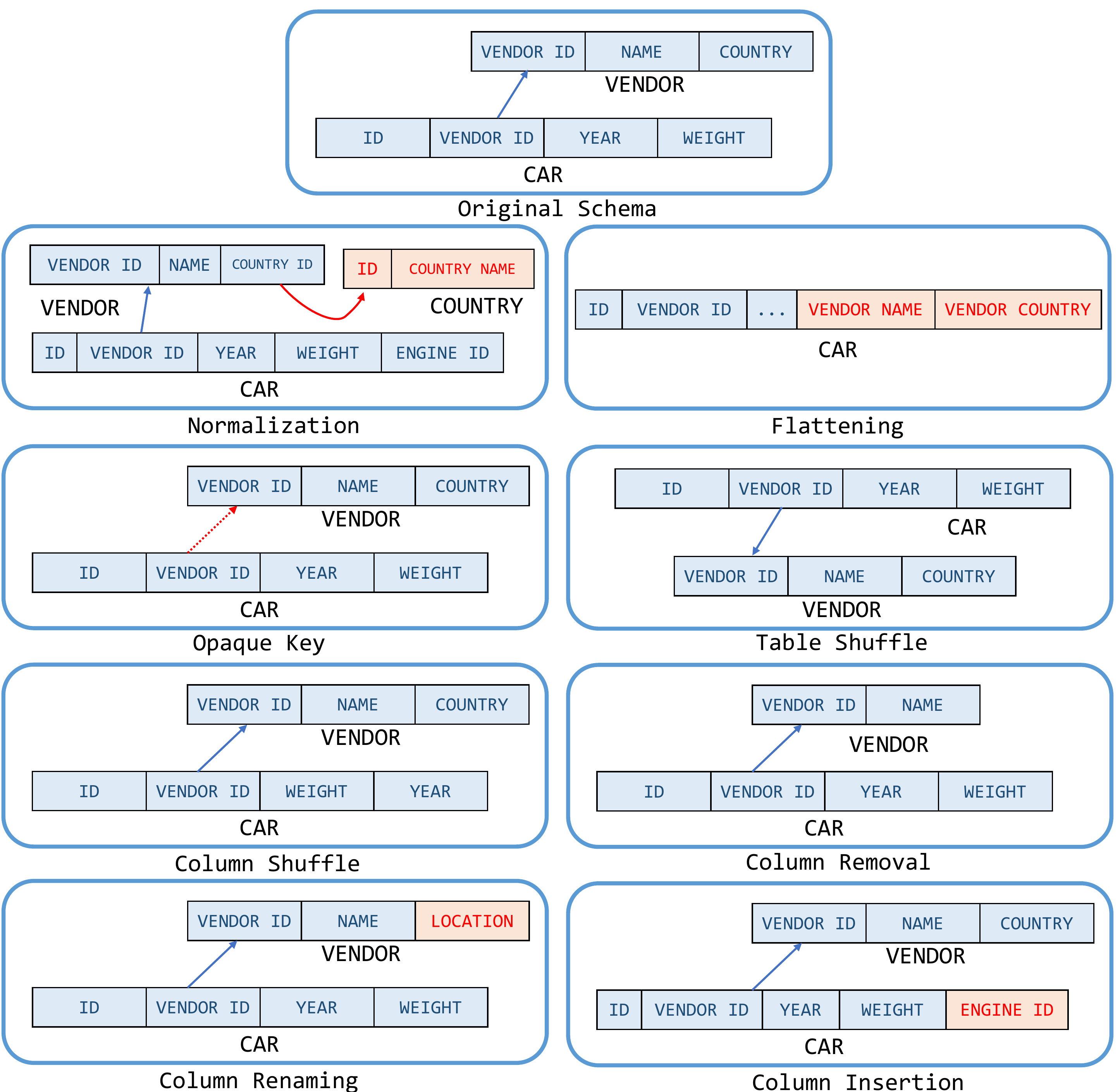}
\caption{Illustrative examples of our designed MRs on schemas.}
\label{fig:schema-mutation}
\end{figure}

As aforementioned, existing works generally test/augment models by transforming
natural language text; nevertheless, given that utterances and database schemas
are learnt jointly by typical text-to-SQL models, we see it as demanding to
further transform database schemas to comprehensively explore model defects. At
this step, we design eight MRs to conduct semantics-preserving transformations
on schemas. We also note that, to preserve output invariance, \teql\ is
carefully designed to only transform a part of schemas which does \textit{not}
change the ground-truth query (e.g., performing extra table joining).
Illustrative examples of the schema-oriented MRs are shown in
\F~\ref{fig:schema-mutation}.

From a holistic view, on the schema structural level, users may decide to
normalize some inter-dependent column or denormalize two tables to boost query
performance. In a more realistic setting, users may not explicitly declare
foreign key constraints or primary key constraints to reduce database
computational overheads. Also, the order of tables stored in the database should
not affect the output, nor does the order of columns in the table. These
observations motivate the design of the following schema-level transformations.

\smallskip
\noindent \textbf{Normalization.}~Conducting normalization on columns that are
irrelevant to the associated utterance should not change the model output. In
principle, we can employ well-established functional dependency mining
algorithms on the raw table content and normalize dependent columns accordingly.
However, these heavy-weight algorithms are unaffordable when we are mutating
considerable amount of tables to test text-to-SQL models. Instead, as shown in
\F~\ref{fig:schema-mutation}, \teql\ launches a light-weight approach to
extracting one unused column each time to form a new table with two columns (one
original column and one linking column). It then links the original table to the
new table with the linking table.

\smallskip
\noindent \textbf{Flattening.}~As a dual operation to normalization, we also
implement MRs to flatten a table if there exists an explicit foreign key
constraint. To be precise, \teql\ piggybacks the reference table to the main
table and drops the reference table. This way, we effectively changes the schema
structure while retaining the high-level semantics.

\smallskip
\noindent \textbf{Opaque Key.}~Explicit foreign key constraints and primary key
constraints can sometimes give hints for models to perform table joining.
However, in practice, explicit key constraints are not always available due to
various reasons (e.g., performance boosts). Though key constraints can be used
as a strong indicator for table joining, a model is expected to be consistent no
matter an explicit key constraint exists or not, as the absence of key
constraints often occurs in the wild. As shown in \F~\ref{fig:schema-mutation},
the explicit foreign key constraint between \texttt{CAR.VENDOR\_ID} and
\texttt{VENDOR.VENDOR\_ID} is removed and it should still be feasible to infer
their dependency relation based on the column names and table names.

\smallskip
\noindent \textbf{Table Shuffle.}~It is easy to see that how tables are stored
in the database system (reflected by the order of tables in the datasets) should
not induce output changes. As shown in \F~\ref{fig:schema-mutation},
\teql\ implements an MR to randomly shuffle tables within a schema and assert
the output consistency.

\smallskip
\noindent \textbf{Column Shuffle.}~Like table shuffle, how columns are stored in
the database system (reflected by the order of columns in the datasets) should
not induce inconsistent SQL queries. Therefore, \teql\ implements an MR to
randomly shuffle columns within a table.

\smallskip
\noindent \textbf{Column Removal+Renaming.}~The name or occurrence of an
\textit{irrelevant} columns should not incur inconsistent outputs. If one column
is not used in the ground-truth query, \teql\ is implemented to change its name
with some synonyms (e.g., ``country'' $\to$ ``location'' in the ``Column
Renaming'' diagram in \F~\ref{fig:schema-mutation}) or simply drop that
particular column.

\smallskip
\noindent \textbf{Column Insertion.}~By querying the knowledge graph, we may
know what attribute the object indicated by the table may have. As shown in the
``Column Insertion'' diagram in \F~\ref{fig:schema-mutation}, \teql\ extends the
schema by inserting an extra column named \texttt{ENGINE\_ID} in the \texttt{CAR}
table, since querying the knowledge graph with ``car'' returns an attribute of
``engine.''

\subsection{Augmentation}
\label{subsec:aug}
Our aforementioned MRs can induce a large volume of synthetic utterances and
schemas (60x compared with the original dataset). We also confirm that these
generated synthetic inputs are well formed and valid (see evaluation in
\S~\ref{subsec:naturalness}). Nevertheless, despite this highly promising
result, we note that it is extremely time-consuming, if at all possible, to
directly augment models using such a large amount of synthetic inputs. Hence,
this section proposes three sampling methods to reduce computational overhead
and practically augment text-to-SQL models.

\smallskip
\noindent \textbf{Random Sampling (RS).}~Suppose we have $m$ synthetic test
cases derived from the training dataset of $n$ samples (where $m \gg n$), our
first sampling strategy, random sampling (RS), randomly picks $n$ test cases
from those $m$ synthetic data. We then extend the training set of $n$ original
samples with those $n$ randomly-picked test cases. This way, the training cost
should not be notably increased by only doubling the size of the training set.

\smallskip
\noindent \textbf{Stratified Sampling (SS).}~Since $m \gg n$, the RS strategy
indeed picks a relatively small portion ($n$ test cases) of the synthetic data
set. In other words, RS may fail to retain inputs generated by certain MRs if
such MRs are applicable to only small amount of data (see
\T~\ref{tab:testcase-dist}).
To retain reasonable amount of data samples for each MR, we further propose the
Stratified Sampling (SS) scheme. In particular, we first sample $\min(m_i,n/k)$
test cases using each MR, where $m_i$ is the total number of test cases
synthesized using this MR and $k$ is the number of MRs we have ($k$ is $12$
according to \T~\ref{tab:strategies}). In case $\sum_{i=1,\cdots,k}
\min(m_i,n/k) < n$, we further randomly sample $n - \sum_{i=1,\cdots,k}
\min(m_i,n/k)$ inputs from the synthetic test cases. This way, we use the SS
scheme to prepare $n$ synthetic test cases where the ``minor'' MRs are guaranteed to
contribute all of its synthesized $m_i$ test cases. We then extend the training
set of $n$ original samples with those $n$ synthetic cases for model
augmentation.

\smallskip
\noindent \textbf{Adaptive Sampling (AS).}~Given that text-to-SQL models can
exhibit inconsistency errors, one might wonder the feasibility of directly using
the error-triggering inputs to augment the dataset. However, we note that the
total amount of error-triggering inputs are \textit{not} directly comparable to
the size of the standard training dataset.
Hence, augmenting the training data with only error-triggering inputs are
not realistic.

Instead, we design Adaptive Sampling (AS), as a practical \textit{error-aware}
sampling scheme.
In particular, we first randomly split the standard training set into ten folds
forming a nine-folds training set $S_t$ and an one-fold validation set $S_v$. A
model $m_0$ is then trained on $S_t$ using half of the standard model training
epoch setting. Then, we transform test cases in $S_v$ using our MRs, and
evaluate $m_0$ in terms of its inconsistency rate $r_i$ (see for
\S~\ref{subsec:setup} the definition of $r_i$) using the synthetic test cases
generated by each MR. We then normalize $r_i$ to $\hat{r_i}$ such that $\sum
\hat{r_i}=1$. Then, similar to SS, we sample $\min(m_i, \hat{r_i}n)$ test cases
from the synthetic data set generated by each MR and further sample from the
remaining test cases to obtain a total of $n$ sampled cases. We train the model
with $n$ original test cases and those $n$ test cases sampled under the
awareness of inconsistency errors.

%% file: tab/mutation-strategies.tex
\begin{table}[t]
  \centering
  \small
		\begin{tabular}{l|c}
      \hline
			\textbf{Target} & \textbf{Metamorphic Relation}\\
      \hline
      \multirow{4}{*}{ Utterance}
                                & Prefix Insertion\\
                                & Prefix Removal \\
                                & Prefix Substitution\\
                                & Synonym Substitution\\
      \hline
      \multirow{8}{*}{ Schema }
                                 & Normalization\\
                                 & Flattening \\
                                 & Opaque Key \\
                                 & Table Shuffle  \\
                                 & Column Shuffle  \\
                                 & Column Removal \\
                                 & Column Renaming  \\
                                 & Column Insertion\\
      \hline
		\end{tabular}
  \caption{Metamorphic relations in \teql.}
  \label{tab:strategies}
\end{table}

%% file: 3-testing.tex
\input{tab/testing-result.tex}

\section{Testing Models with \teql}
\label{sec:test}
Before testing text-to-SQL models with \teql, we first synthesize a total of
62,430 test cases from 1,034 data samples in the Spider dev set~\cite{Spider}.
We report the distribution of test cases in terms of MRs in
\T~\ref{tab:testcase-dist}. A reasonable number of test cases can be synthesized
by all MRs. We note that, as some MRs may have exponential test cases w.r.t.~the
input size (e.g., column shuffle), \teql\ only picks maximally ten possible
candidates derived from one test case for each MR to avoid overwhelming test
cases.

\subsection{Naturalness of Synthetic Utterances}
\label{subsec:naturalness}
As previously mentioned, instead of randomly transforming utterances (e.g.,
replacing certain words with typos) which likely induces ill-formed utterances,
we aim to synthesize \textit{natural} utterances whose incurred inconsistency
likely denotes real-world defects that normal users can encounter. That is, we
advocate the naturalness as an important metric to assess the quality of
synthetic utterances. As there are no well-established metrics to quantify
naturalness, we employ the fluency score~\cite{ge2018fluency} to practically
approximate the naturalness of utterances, which is defined as follows:

\begin{equation}
  f(\boldsymbol{x})=\frac{1}{1+H(\boldsymbol{x})}
\end{equation}

\noindent where $H(\boldsymbol{x})$ is

\begin{equation}
  H(\boldsymbol{x})=-\frac{\sum_{i=1}^{|\boldsymbol{x}|} \log P\left(x_{i} 
  \mid \boldsymbol{x}_{<i}\right)}{|\boldsymbol{x}|}
\end{equation}

\noindent $P\left(x_{i} \mid \boldsymbol{x}_{<i}\right)$ is the
probability of $x_i$ under given context $\boldsymbol{x}_{<i}$. We report that
the average fluency score on the original utterances is $0.155$ while the
average fluency score of synthetic utterances is $0.148$ ($-4.5\%$). It shows
that the synthetic utterances exhibit comparable fluency with the original
utterances.

\begin{figure}[!htbp]
\centering
\includegraphics[width=\columnwidth]{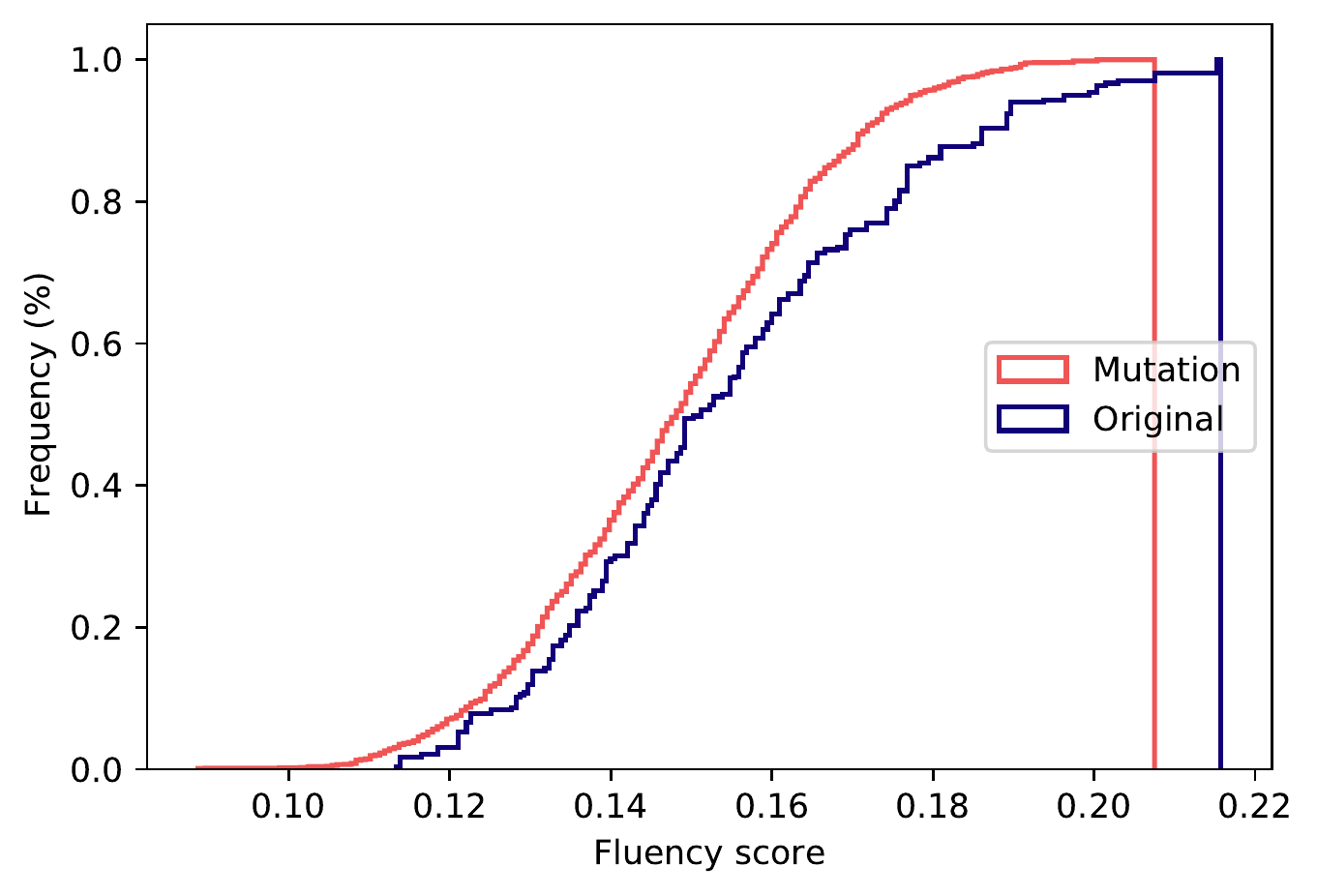}
\caption{Cumulative fluency score distributions.}
\label{fig:cdf}
\end{figure}

\F~\ref{fig:cdf} further reports the cumulative fluency scores distributions of
original utterances and synthetic utterances. Particularly, in bottom 25\%, the
fluency score distributions of original utterances and synthetic utterances are
very close, illustrating promising results that our MRs do not notably impede
the readability of worse-case utterances in the original dataset.

\subsection{Setup}
\label{subsec:setup}
As summarized in \T~\ref{tab:testing-model-info}, we reproduce six models and
test them with \teql.
The tested models feature diverse human utterance embedding techniques as well
as schema learning modules, which, as will be shown in \S~\ref{subsec:findings},
incur different model consistency w.r.t.~different MRs. We also report that,
though we put much effort to resolve incompatibility issues (e.g., fixing buggy
data reader and query parser) of the official model implementations, some models
still fail to process a small proportion ($\approx 3\%$ on average) of our
synthetic test cases. We ignore these test cases in the evaluation.

\noindent \textbf{Inconsistency Rate.}~We first propose the formulation of
inconsistency rate as a quantifier of the model robustness. Given a test suite
$\mathcal{S}_i=\{((u_1,s_1),(u_{11},s_{11})),\cdots,
((u_n,s_n),(u_{nm},s_{nm}))\}$, where $(u_i,s_i)$ denotes the $i$-th sample in
the original dataset and $(u_{ij},s_{ij})$ denotes the $j$-th test case derived
from $(u_i,s_i)$, we define the inconsistency rate as follows:

\begin{equation}
    r_i=\frac{\sum \mathbb{1}_{eval(\mathcal{M}(u_i,s_i),\mathcal{M}(u_{ij},s_{ij}))}}{|\mathcal{S}_i|}
\end{equation}

\noindent where $\mathbb{1}_{eval(\mathcal{M}(u_i,s_i),\mathcal{M}(u_{ij},s_{ij}))}$ 
is an indicator function and it returns $1$
if the model $\mathcal{M}$ yields \textit{inconsistent} outputs over $(u_i,s_i)$
and $(u_{ij},s_{ij})$. Here, $eval(\cdot,\cdot)$ can be any standard evaluation
metrics, for instance, exact set match (EM)~\cite{Spider} or semantics
equivalence~\cite{chu2017cosette}. At this step, we employ exact set match (EM)
accuracy, the most widely-used metric, as our $eval(\cdot,\cdot)$.

\begin{figure*}[!htbp]
\centering
\includegraphics[width=\textwidth]{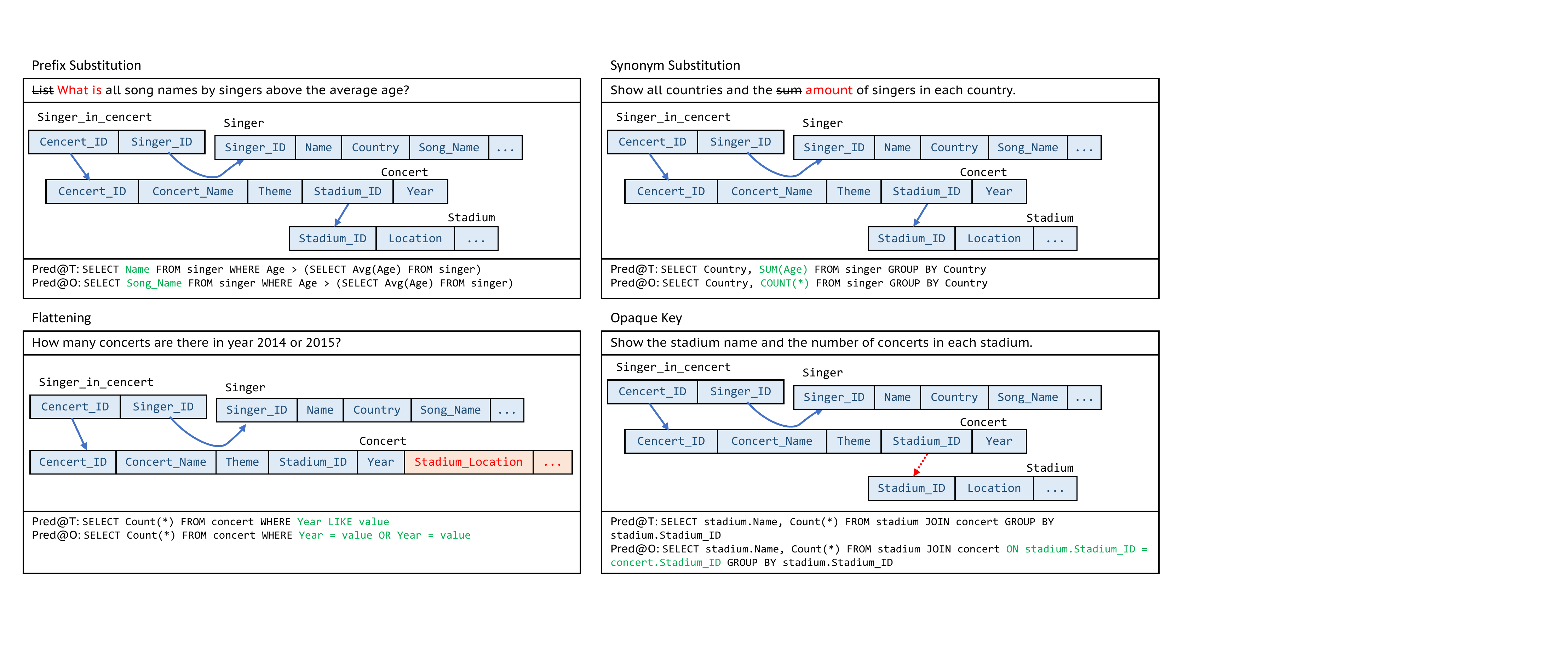}
\caption{Cases of inconsistency errors. Foreign key constraints are annotated 
by arrows; transformations are highlighted in {\color{red}red}; inconsistencies
are highlighted in {\color{green}green}; \texttt{Pred@T} denotes the prediction
on the transformed inputs and \texttt{Pred@O} denotes the prediction on the
original inputs.}
\label{fig:case}
\end{figure*}

\begin{table}[!htbp]
    \centering
    \resizebox{\linewidth}{!}{
		\begin{tabular}{l|c|c|c|c|c}
      \hline
			\textbf{Model} & \textbf{Easy} &\textbf{Medium}&\textbf{Hard}&\textbf{Extra}&\textbf{All}\\
      \hline
      SyntaxSQLNet & 5.2& 3.7& 1.4&0.8&3.2\\\hline
      IRNet & 2.1& 4.7& 2.1& 6.9&4.0\\\hline
      GNN & 1.9& 2.5& 2.4& 2.6& 2.4\\\hline
      GlobalGNN+Linking & 2.8& 4.0& 2.0& 3.1&3.2\\\hline
      RAT-SQL (v1) & 1.2& 2.4& 1.5& 1.7&1.9\\\hline
		\end{tabular}}
  \caption{Testing results of models under \teql\ (breakdown by hardness, measured by 
  inconsistency rate).}
  \label{tab:testing-rlt-hardness}
\end{table}

\subsection{Findings}
\label{subsec:findings}
We report the inconsistency rate of models in \T~\ref{tab:testing-rlt-mr} and
\T~\ref{tab:testing-rlt-hardness}. \teql\ successfully finds
considerable bugs from all the evaluated text-to-SQL models.
\T~\ref{tab:testing-rlt-mr} reports the evaluation results in terms of different
MRs. We interpret that utterance transformations are all highly
effective to expose inconsistent outputs. Synonym Substitution (\textbf{SS})
is particularly effective, given its high flexibility in perturbing natural
language utterances. Relatively mundane prefix-oriented transformations, which
impose easy challenge for human, are also very effective to stress text-to-SQL
models.

Schema-based transformations also achieve reasonable performance in exposing
model defects. We note that the consistency of models against schema-based
transformations is highly influenced by its schema learning module
(see \T~\ref{tab:testing-model-info}). For example, SyntaxSQLNet and IRNet use
the sequences of columns as input and are generally more sensitive to Table
Shuffle (\textbf{TS}) and Column Shuffle (\textbf{CS}). In contrast, the other
models learn from schema structural feature (e.g., graph-based method and
relation encoding) and become more resilient against such transformations.
Moreover, more holistic structure changes, e.g., Normalization (\textbf{NO}) and
Flattening (\textbf{FL}), impose noticeable challenges to these models, which
are intuitive.

\T~\ref{tab:testing-rlt-hardness} further reports the inconsistency rates
breakdown by the hardness. Overall, \teql\ effectively reveals subtle
differences in the predictions of de facto models. Models of different
architectures and data preprocessing show largely distinct robustness against
questions of different hardness. For instance, SyntaxSQLNet performs apparently
worse regarding ``Easy'' questions, while it shows high consistency regarding
``Hard'' and ``Extra'' questions. Overall, we view \T~\ref{tab:testing-rlt-mr}
and \T~\ref{tab:testing-rlt-hardness} have clearly illustrated the strength of
our proposed approach, which can expose considerable defects from well-trained
de facto models. Furthermore, as a by-product, \teql\ can serve as an
\textit{assessment criteria} to help engineers make fine-grained diagnosis 
on their text-to-SQL models, exposing their subtle difference and preference.

\begin{table}[!htbp]
  \centering
  \scriptsize
  \begin{tabular}{l|c}
    \hline
    \textbf{Type} & \textbf{\# Errors}\\
    \hline
    Column Prediction & 101\\\hline
    Aggregate Function & 77\\\hline
    Operator & 73\\\hline
    Table Joining & 171\\\hline
    Complex Query & 48\\\hline
    Others & 30\\\hline
  \end{tabular}
\caption{Error type distributions of 500 inconsistency errors.}
\label{tab:error-dist}
\end{table}

\subsection{Case Study and Error Analysis}
\label{subsec:case}
To understand the details of inconsistency errors, we report four representative
cases of inconsistency errors in \F~\ref{fig:case}. They are derived by
transforming the \texttt{Singer} schema in the Spider dev set or transforming
utterances associated with the schema. We present the SQL query derived from the
original inputs (\texttt{Pred@O}) and the query derived from the transformed
inputs (\texttt{Pred@T}), accordingly.

We manually checked 500 pairs of inconsistency errors and find that most errors
have correct query skeletons and incorrect or missing clauses. We also
notice that there is no false negative cases in the inconsistency errors and
presume the overall false negative should be lower than error cases reported by
standard metrics. For example, it is reported that exact set matching has a
2.5\% false negative rate on average and 8.1\% in the worst
case~\cite{zhong2020semantic}. We hypothesis that, given two similar inputs
(i.e., an original input and a transformed input), models are not likely to
yield two semantically equivalent while syntactically different outputs.
Therefore, we envision it as a useful by-product of MT compared with standard
methods. Following, we discuss five major sources of inconsistency errors.

\smallskip
\noindent \textbf{Column Predictions.}~As shown in the ``Prefix 
Substitution'' of \F~\ref{fig:case}, the model is misled to choose an incorrect
column \texttt{Name} instead of \texttt{Song\_Name}, even though the correct
column name is explicitly referred to in the utterance. In contrast, as reported
by the error analysis on the standard dataset~\cite{guo2019towards}, most column
prediction errors are due to that the ground-truth column names are not
explicitly (or merely partially) mentioned in the utterances. According to our
observation, column prediction errors not only exist in the \texttt{SELECT}
clause, but also occur in the \texttt{GROUP BY} and \texttt{ORDER BY} clauses.
According to our manual study of 500 errors, erroneous column prediction account
for 20.2\% errors.

\smallskip
\noindent \textbf{Aggregate Function.}~Among 500 manually checked
errors, we report that 15.4\% errors are caused by incorrect aggregate function.
As discussed in
\S~\ref{subsec:mutation}, changing the aggregate function indicator can
impose challenges to the text-to-SQL models. For instance, in the ``Synonym
Substitution'' case of \F~\ref{fig:case}, by replacing ``sum'' with ``amount
of'', the tested model fails to comprehend the transformed utterance and uses an
incorrect aggregate function \texttt{SUM(age)} in \texttt{Pred@T}, although
``age'' was never mentioned in the original and the transformed utterances.

\smallskip
\noindent \textbf{Operator.}~Predictions on operator is a major step 
towards identifying intended contents. Our manual study on the 500 erroneous
cases report that 14.6\% errors are due to incorrect operator predictions. For
example, a \texttt{LIKE} operator may be mistakenly predicted as
\texttt{=}. As shown in the ``Flattening'' erroneous case of \F~\ref{fig:case}, 
some irreverent changes on schema structure can result in an incorrect prediction 
on the operator in the \texttt{WHERE} clause. Besides, we also observe errors 
in the \texttt{HAVING} clauses of the group-by operation. Overall, our manual 
study shows that such erroneous operators widely exist in almost all MRs.

\smallskip
\noindent \textbf{Table Joining.}~Linking multiple tables in one query is 
challenging for models and is a major cause of inconsistency errors (34.2\%).
As an important table joining indicator, foreign key constraints help models 
link two tables to a large degree. Without explicit foreign key constraints in 
the schema, models may lack of guidance to link two tables. As shown in the 
``Opaque Key'' case of \F~\ref{fig:case}, though the model captures the need 
of table joining, it fails to make prediction on the joining condition 
(i.e., the \texttt{ON} clauses). Besides ``Opaque Key'', our manual study shows 
that prefix- and column-based transformations can also trigger a considerable number 
of table joining errors.

\smallskip
\noindent \textbf{Complex Query.}~In addition to the four basic types of 
inconsistency errors, we also observe that some ``extra hard'' queries in the
Spider dev set, with multiple table joining and complex conditions, are itself
hard for models, even for human experts, to translate. Hence, even subtle
changes on the input can induce multiple errors in the corresponding output.
Nevertheless, as will be reported in \S~\ref{sec:aug}, dataset augmented by our
transformed inputs can effectively enhance model performance toward ``extra
hard'' challenges. Among the sampled 500 pairs, we observe that about 9.6\% 
predictions have more than one errors.

\smallskip
\noindent \textbf{Discussion.}~We have shown common weaknesses of the
text-to-SQL models. Besides the mentioned types of errors, we also find some 
errors are less common, e.g., erroneous table predictions and unwanted
conditions in \texttt{WHERE} clauses, which consist 6.0\% of sampled errors.
On the other hand, we find that substantial errors can be
potentially eliminated with human annotation or interactive correction. For
example, if users can explicitly specify their desired aggregate functions or
interested columns in addition to the input utterances, the search space of the
model can be effectively reduced, presumably enhancing the model accuracy.
Hence, we envision adopting human-in-the-loop methods to further disambiguate
user intents and reduce the model search space (e.g., by dropping irreverent
columns and tables)~\cite{andreas2020task,li2020you}.
From the model design side, we have observed considerable potential improvements
with more accurate natural language understanding. The model can generally yield
more consistent outputs when processing utterances of different prefixes. In
that sense, we presume task-oriented sentence representation learning techniques
can facilitate natural language comprehension toward even noisy
utterances~\cite{yu2020grappa}. In terms of schema variations, we observe that 
graph neural networks have better robustness under subtle changes on schemas while become
even more vulnerable on holistic structural changes. In general, it is hard for neural 
network models to maintain consistent performance under noisy schemas~\cite{suhr2020exploring}. 
We observe that augmenting standard dataset with more schema variations can 
boost the model robustness, as will be shown shortly in \S~\ref{sec:aug}.

%% file: tab/testing-result.tex
\begin{table}[!ht]
  \centering
  \small
		\begin{tabular}{l|c}
      \hline
			\textbf{Metamorphic Relation} & \textbf{\# Test Case}\\
      \hline
        Prefix Insertion & 6370\\
        \hline
        Prefix Removal & 199\\
        \hline
        Prefix Substitution & 8266\\
        \hline
        Synonym Substitution & 639\\
      \hline
        Normalization & 8707\\
        \hline
        Flattening & 2018\\
        \hline
        Opaque Key & 3697\\
        \hline
        Table Shuffle & 2575\\
        \hline
        Column Shuffle  & 6675\\
        \hline
        Column Removal & 8707\\
        \hline
        Column Renaming & 11775 \\
        \hline
        Column Insertion & 2802\\
      \hline
        \textbf{Total} & 62,430 \\
      \hline
		\end{tabular}
  \caption{Distribution of generated test cases in terms of each MR.}
  \label{tab:testcase-dist}
\end{table}

\begin{table*}[!htbp]
  \centering
  \resizebox{\linewidth}{!}{
  \begin{tabular}{l|c|c|c|c}
    \hline
    \textbf{Model} & \textbf{Year}&\textbf{Utterance Encoder} & \textbf{Schema Learning Module} &\textbf{Accuracy}\\
    \hline
    SyntaxSQLNet~\cite{yu2018syntaxsqlnet} & 2018& GloVe~\cite{pennington2014glove} & Column Sequence&22.4\\\hline
    IRNet (v1)~\cite{guo2019towards} & 2019& GloVe & Column Sequence+Linking& 52.8\\\hline
    GNN~\cite{bogin-etal-2019-representing} & 2019& Bi-LSTM & Schema Graph&45.0\\\hline
    GlobalGNN+Linking~\cite{bogin2019global,chen2020tale} & 2020& BERT-base~\cite{devlin2019bert} & Schema Graph+Linking&54.9\\\hline
    RAT-SQL (v1)~\cite{wang2019rat} & 2019& GloVe& Relation Encoding+Linking&53.5\\\hline
  \end{tabular}}
\caption{Information of our reproduced models.}
\label{tab:testing-model-info}
\end{table*}

\begin{table*}[t]
    \centering
      \small
		\begin{tabular}{l|c|c|c|c|c|c|c|c|c|c|c|c|c}
      \hline
			\textbf{Model} & \textbf{PI} &\textbf{PR}&\textbf{PS}&\textbf{SS}&\textbf{NO}&\textbf{FL}&\textbf{OK}&\textbf{TS}&\textbf{CS}&\textbf{CRm}&\textbf{CRn}&\textbf{CI}&\textbf{ALL}\\
      \hline
      SyntaxSQLNet & 8.2& 8.0& 6.7&13.3& 2.4&2.7&1.3&2.1&1.1&1.5& 1.7&0.8&3.2\\\hline
      IRNet & 7.5& 11.2& 7.0& 37.8& 2.4&6.6&5.2&2.8&2.1& 2.1&1.6&2.1&4.0\\\hline
      GNN & 4.6& 5.5& 3.4& 25.4& 3.4& 4.3& 5.3& 0.0& 0.0& 1.0& 0.5& 0.2& 2.4\\\hline
      GlobalGNN+Linking & 4.0& 7.5& 3.3& 18.3& 5.8&6.9&6.1&0.4&0.2&2.9& 1.1&3.0&3.2\\\hline
      RAT-SQL (v1)& 2.4& 5.0& 2.6& 21.8& 3.0& 4.9& 3.8& 0.2&0.04&1.3& 0.7&0.0&1.9\\\hline
      \textbf{Average} & 5.3& 7.4& 4.6& 23.3& 3.4& 5.1& 4.3& 1.1& 0.7& 1.8& 1.1& 1.2& 2.9\\\hline
		\end{tabular}
  \caption{Testing results of models under \teql\ (breakdown by MRs, measured by
    inconsistency rate). \textbf{PI}: Prefix Insertion; \textbf{PR}: Prefix
    Removal; \textbf{PS}: Prefix Substitution; \textbf{SS}: Synonym
    Substitution; \textbf{NO}: Normalization; \textbf{FL}: Flattening;
    \textbf{OK}: Opaque Key; \textbf{TS}: Table Shuffle; \textbf{CS}: Column
    Shuffle; \textbf{CRm}: Column Removal; \textbf{CRn}: Column Renaming;
    \textbf{CI}: Column Insertion.}
  \label{tab:testing-rlt-mr}
\end{table*}

%% file: 4-augmenting.tex

\input{tab/augmenting-result.tex}

\section{\teql\ for Augmentation}
\label{sec:aug}

In addition to model testing, we also seek to augment models with moderate cost,
as discussed in \S~\ref{subsec:aug}. To evaluate the proposed method, we perform
three proposed sampling techniques on SyntaxSQLNet and IRNet. We also compare
our method with the SyntaxSQLNet data augmentation module.

To present a fair comparison, all variants of SyntaxSQLNet are trained with 600
epochs on \texttt{col} model and 300 epochs on the remaining sub-models; all
variants of IRNet are trained with 50 epochs. Since the standard data
augmentation module in SyntaxSQLNet uses 3x more data samples from external
dataset, we additional design the ``AS$^+$'' group that samples more synthetic
data and makes the size of its training set equal to ``aug'' for a fair
comparison.

We report the inconsistency rates in \T~\ref{tab:augmentation-rlt-mr}, subsuming
the original model and also the augmented models with three schemes.
Furthermore, we also report the accuracy of (augmented) models in
\T~\ref{tab:augmentation-rlt}. Overall, \T~\ref{tab:augmentation-rlt-mr} shows
that with augmentation, the model inconsistency is notably reduced. Consistent
with our intuition, the adaptive sampling (AS) scheme (and AS$^{+}$), which
entails an ``error-aware'' augmentation, exhibits the best performance to reduce
model prediction inconsistency.

\T~\ref{tab:augmentation-rlt} shows that the overall model performance 
is slightly reduced. On one hand, it is generally acknowledged that models with
relatively higher robustness can exhibit lower accuracy to particular training
data. In particular, \T~\ref{tab:augmentation-rlt-mr} has shown that
SyntaxSQLNet, with being augmented by previous work (the ``SyntaxSQLNet+aug''
row), becomes less robust with an average inconsistency rate of 3.4. On the
other hand, we would like to point out that de facto models, with being augmented by
the AS$^{+}$ scheme, performs generally comparable in terms of ``Hard'' and
``Extra'' questions with ``SyntaxSQLNet+aug'' (see the last column of
\T~\ref{tab:augmentation-rlt}), while achieving much better robustness
as shown in \T~\ref{tab:augmentation-rlt-mr}.


%% file: tab/augmenting-result.tex
\begin{table*}[!htbp]
    \centering
      \small
		\begin{tabular}{l|c|c|c|c|c|c|c|c|c|c|c|c|c}
      \hline
			\textbf{Model} & \textbf{PI} &\textbf{PR}&\textbf{PS}&\textbf{SS}&\textbf{NO}&\textbf{FL}&\textbf{OK}&\textbf{TS}&\textbf{CS}&\textbf{CRm}&\textbf{CRn}&\textbf{CI}&\textbf{ALL}\\
      \thickhline
      SyntaxSQLNet & 3.7& 8.9& 5.0&22.2& 2.6&8.1&1.3&1.5&0.6&1.5& 1.1&0.0&2.5\\\hline
      SyntaxSQLNet+aug& 9.2& 3.5& 8.2&16.7& 2.1&2.2&1.4&2.3&1.1&1.4& 1.6&1.1&3.4\\\hline
      SyntaxSQLNet+RS & 5.2&4.0 & 5.1& 8.1& 1.2& 2.2& 0.9&1.6&0.5&0.9& 1.0&0.3&2.0\\\hline
      SyntaxSQLNet+SS & 4.7& 3.5& 4.6& 7.8& 0.9& 2.0&0.8&1.4&0.3&0.5& 0.9&0.4&1.8\\\hline
      SyntaxSQLNet+AS & 3.9& 6.5& 4.3&7.2& 0.8&1.4& 0.3& 1.4& 0.2&0.6& 0.8&0.9&1.6\\\hline
      SyntaxSQLNet+AS$^+$ & 3.5& 4.5& 3.8&2.2& 1.5&2.4& 1.1& 1.8& 0.8&1.1& 0.9&0.2&1.8\\\thickhline
      IRNet & 7.5& 11.2& 7.0& 37.8& 2.4&6.6&5.2&2.8&2.1& 2.1&1.6&2.1&4.0\\\hline
      IRNet+RS & 4.3& 7.6& 4.7& 14.9& 1.0&3.4& 2.8& 0.5&0.5&0.9& 0.5&0.1&1.9\\\hline
      IRNet+SS & 5.2& 8.1& 5.5&17.3& 1.2& 4.7& 3.6& 1.1&0.7&1.0&0.8& 0.3& 2.4\\\hline
      IRNet+AS & 4.6& 8.1& 5.7& 10.2& 1.0&2.5&3.0&0.6&0.4&0.7& 0.9&0.1&2.1\\\hline
		\end{tabular}
  \caption{Inconsistency rate of augmented models. \textbf{aug}: SyntaxSQLNet standard data augmentation;
  \textbf{RS}: Random Sampling; \textbf{SS}: Stratified Sampling; \textbf{AS}: Adaptive Sampling.}
  \label{tab:augmentation-rlt-mr}
\end{table*}

\begin{table}[!htbp]
  \centering
  \resizebox{\linewidth}{!}{
  \begin{tabular}{l|c|c|c|c|c|c}
      \hline
    \textbf{Model} & \textbf{Easy} &\textbf{Medium}&\textbf{Hard}&\textbf{Extra}&\textbf{All} &\textbf{All$^{*}$}\\
    \thickhline
    SyntaxSQLNet & 41.9& 19.3& 17.8&6.6&22.4     & 12.4 \\\hline
    SyntaxSQLNet+aug & 42.7& 24.2& 22.4&8.4&25.8 & 15.6 \\\hline
    SyntaxSQLNet+RS & 37.9& 20.6& 23.0&6.0&22.8  & 14.7 \\\hline
    SyntaxSQLNet+SS & 41.1& 18.8& 20.1&9.0&22.8  & 14.7 \\\hline
    SyntaxSQLNet+AS & 37.5& 18.4& 19.5&7.2&21.4  & 13.5 \\\hline
    SyntaxSQLNet+AS$^+$ & 42.3& 21.7& 21.3&8.4&24.5  & 15.0 \\\thickhline
    IRNet & 72.3& 53.6& 42.0& 32.9&52.8          & 37.6 \\\hline
    IRNet+RS & 70.3& 52.9& 47.7& 30.5&52.6       & 39.3 \\\hline
    IRNet+SS & 70.7& 53.8& 45.9& 35.9&52.7       & 41.1 \\\hline
    IRNet+AS & 70.3& 54.3& 44.8& 29.3&52.5              & 37.3 \\\hline
    \end{tabular}}
\caption{Standard accuracy results of augmented models (breakdown by hardness, measured by 
exact set matching rate \cite{Spider}).}
\label{tab:augmentation-rlt}
\end{table}

%% file: 7-related.tex
\section{Related Work}
\label{sec:related}

\noindent \textbf{Text-to-SQL and generalizability.}~Text-to-SQL (or NL2SQL) 
is a challenging topic which requires to comprehend and translate human 
utterances into corresponding structured SQL queries toward the given 
relational database. It has been actively studied by both NLP and database 
communities for decades~\cite{kim2020natural}. Recently, AI-powered models 
manifest highly promising performance on cross-domain and cross-table 
text-to-SQL tasks and show great potential for real-world
usage~\cite{zhongSeq2SQL2017,xu2017sqlnet,yu2018typesql,Spider,yu2018syntaxsqlnet,bogin2019global,guo2019towards,wang2019rat,chen2020tale}.
The generalizability of these models on unseen user cases is critical for 
commercial adaption but not systematically evaluated. Suhr et al.~pinpoint
the main challenge on linguistic variations, novel database and query structure
and conventions in different datasets \cite{suhr2020exploring}.
Data augmentation techniques are also extensively used to enrich utterances
and achieve higher performance on standard
metrics~\cite{yu2018syntaxsqlnet,yu2020grappa,weir2020dbpal}. However, given 
the unbalanced size of schemas and utterances, e.g., 166 schemas vs.~9693 
utterances in the Spider dataset~\cite{Spider}), it is indeed necessary to 
further augment schemas in the standard dataset. In addition, given the
variations of human languages, it may be difficult for existing template- or
grammar-based augmentation methods to further enhance consistency w.r.t.~noisy
utterances in the wild. 

\smallskip
\noindent \textbf{Robustness of NLP Models and Model Testing.}~Adversarial
examples, by substituting original tokens in an utterance with typos or
synonyms, impede model performance deliberately~\cite{zhang2020adversarial}.
Photon augments text-to-SQL models with untranslatable user inputs to improve
the model robustness~\cite{zeng2020photon}. Furthermore, instead of evaluating
adversarial robustness, some works advocate a focus on assessing models with
software testing methods and expose prediction
inconsistency~\cite{ma2020metamorphic,he2020structure,ribeirobeyond}. By mostly
preserving the ``naturalness'' of mutated inputs, their findings would
presumably indicate defects and confusions that normal users can encounter in
real-life scenarios. Compared with this line of work, this present research
takes one major step further by systematically transforming both natural
language texts and database schemas. We also propose practical (error-aware)
data augmentation strategies using the mutated inputs to enhance model
robustness.
